\documentclass[letterpaper, 10 pt, conference]{cls/ieeeconf}
\IEEEoverridecommandlockouts
\let\labelindent\relax

\renewcommand{\baselinestretch}{0.96} 

\usepackage[normalem]{ulem}                         
\usepackage[table,usenames,dvipsnames]{xcolor}      
\usepackage{enumitem}                               
\usepackage{extarrows}                              
\usepackage[noadjust]{cite}                         
\usepackage{verbatim}

\usepackage{amsmath,amssymb,amsfonts,dsfont} 
\usepackage{amsthm}
\usepackage{bbm}
\usepackage{algorithm,algorithmicx,listings}        
\usepackage[noend]{algpseudocode}             

\usepackage{graphicx,tabularx,subcaption}
\usepackage[export]{adjustbox}
\usepackage{makecell,booktabs}

\usepackage[font={small}]{caption}
\usepackage[titletoc,title]{appendix}

\usepackage[breaklinks=true, colorlinks, bookmarks=true, citecolor=Black, urlcolor=Violet,linkcolor=Black]{hyperref}

\newcommand{\indicator}{\mathds{1}}

\newcommand{\scaleMathLine}[2][1]{\resizebox{#1\linewidth}{!}{$\displaystyle{#2}$}}
\newcommand{\prl}[1]{\left(#1\right)}

\newcommand{\crl}[1]{\left\{#1\right\}}

\usepackage{mathtools}
\usepackage{pifont}
\usepackage{footnote}
\usepackage{tikz}
\makesavenoteenv{tabular}
\makesavenoteenv{table}
\newcommand{\defeq}{\vcentcolon=}

\def\etal/{et~al.}
\newcommand{\cmark}{{\color{green}\ding{51}}}%
\newcommand{\xmark}{\color{red}\ding{55}}%
\graphicspath{{fig/}}

\newtheorem{proposition}{Proposition}

\newtheorem{lemma}{Lemma}
\theoremstyle{definition}

\newtheorem*{assumption*}{Assumption}
\newtheorem*{problem*}{Problem}

\theoremstyle{remark}

\newtheorem*{solution*}{Solution}

\def\thetitle{Learning Navigation Costs from Demonstration in Partially Observable Environments}
\def\theauthor{Tianyu Wang, Vikas Dhiman, and Nikolay Atanasov}
\def\thekeywords{keywords}

\hypersetup{
  pdfauthor={\theauthor},%
  pdftitle={\thetitle},%
  pdfsubject={\thetitle},%
  pdfkeywords={\thekeywords}
}

\newcommand{\calL}{{\cal L}}

\newcommand{\calT}{{\cal T}}
\newcommand{\calU}{{\cal U}}

\newcommand{\calX}{{\cal X}}

\newcommand{\bfg}{\mathbf{g}}
\newcommand{\bfh}{\mathbf{h}}

\newcommand{\bfm}{\mathbf{m}}

\newcommand{\bfu}{\mathbf{u}}

\newcommand{\bfx}{\mathbf{x}}
\newcommand{\bfy}{\mathbf{y}}
\newcommand{\bfz}{\mathbf{z}}

\newcommand{\bftheta}{\boldsymbol{\theta}}

\newcommand{\bftau}{\boldsymbol{\tau}}

\newcommand{\bfphi}{\boldsymbol{\phi}}

\newcommand{\bfpsi}{\boldsymbol{\psi}}

\title{\LARGE \bf \thetitle}
\author{Tianyu Wang \and Vikas Dhiman \and Nikolay Atanasov
\thanks{We gratefully acknowledge support from NSF CRII IIS-1755568, ARL DCIST CRA W911NF-17-2-0181, and ONR SAI N00014-18-1-2828.}
\thanks{The authors are with the Department of Electrical and Computer Engineering, University of California, San Diego, La Jolla, CA 92093, USA {\tt\small \{tiw161,vdhiman,natanasov\}@eng.ucsd.edu}}
}

\begin{document}
\maketitle
\thispagestyle{empty}
\pagestyle{empty}


\begin{abstract}
This paper focuses on inverse reinforcement learning (IRL) to enable safe and
efficient autonomous navigation in unknown partially observable environments.
The objective is to infer a cost function that explains expert-demonstrated
navigation behavior while relying only on the observations and state-control
trajectory used by the expert.
We develop a cost function representation composed of two parts: a probabilistic
occupancy encoder, with recurrent dependence on the observation sequence, and a
cost encoder, defined over the occupancy features.
The representation parameters are optimized by differentiating the error between 
demonstrated controls and a control policy computed from the cost encoder.
Such differentiation is typically computed by dynamic programming through 
the value function over the whole state space.
We observe that this is inefficient in large partially observable environments
because most states are unexplored.
Instead, we rely on a closed-form subgradient of the cost-to-go obtained only
over a subset of promising states via an efficient motion-planning algorithm
such as A* or RRT.
Our experiments show that our model exceeds the accuracy of baseline IRL algorithms in 
robot navigation tasks, while substantially improving the efficiency of training 
and test-time inference.
\end{abstract}

\section{Introduction}
\label{sec:introduction}



Practical applications of autonomous robot systems increasingly require
operation in unstructured, partially observed, unknown, and changing
environments.
Achieving safe and robust navigation in such conditions is directly coupled with
the quality of the environment representation and the cost function specifying
desirable robot behavior.
Designing a cost function that accurately encodes safety, liveness, and
efficiency requirements of navigation tasks is a major challenge.
In contrast, it is significantly easier to obtain demonstrations of desirable behavior.
The field of inverse reinforcement
learning~\cite{Ng_IRL00,IRLSurvey,Neu_Apprenticeship12} (IRL) provides numerous
tools for learning cost functions from expert demonstration.

We consider the problem of safe navigation from partial observations in 
unknown environments. We assume that demonstrations containing expert poses,
controls, and sensory observations over time are available. The objective is to
infer the cost function, which depends on the observation sequence, and explains
the demonstrated behavior. This inference can only be done indirectly by
comparing the control inputs, that a robot may take based on its current cost representation, to the expert's actions in that situation.

\begin{figure}
  \includegraphics[width=\linewidth]{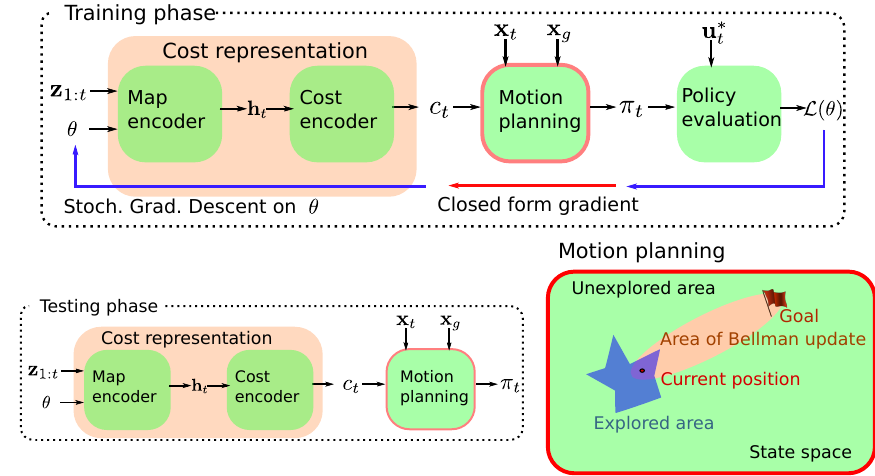}
  \caption{
    Architecture for learning cost function representations from demonstrations. Our main contribution lies in a non-stationary cost representation, combining a probabilistic occupancy \emph{map encoder}, with recurrent dependence on observations $\bfz_{1:t}$, and a \emph{cost encoder}, defined over the occupancy features. We also perform efficient (in the size of the state-space) forward non-stationary policy computation and efficient (closed-form subgradient) backpropagation. The bottom-right plot illustrates that motion planning may be used to update and differentiate cost-to-go estimates only in promising areas of the state space rather than using full Bellman backups, which is particularly redundant in partially observable environments.}
  \label{fig:approach}
\end{figure}

Learning a cost function requires a differentiable control policy 
with respect to the stage cost parameters.
Computing such derivatives has been addressed by several successful
approaches~\cite{Ratliff_06, Ziebart_MaxEnt08, Tamar_VIN16, Okada_PIN17}.
Ratliff~et~al.~\cite{Ratliff_06} developed algorithms with regret bounds for
computing subgradients of planning algorithms (e.g., A*~\cite{ARA},
RRT~\cite{Lavalle_RRT98,Karaman_RRTstar11}, etc.) with respect to the cost
features.
Ziebart~et~al.~\cite{Ziebart_MaxEnt08} developed a dynamic programming algorithm
for computing the expected state visitation frequency of a policy extracted from
expert demonstrations to enforce that it earns the same reward as the
demonstrated policy.
Tamar~et~al.~\cite{Tamar_VIN16} showed that the value iteration algorithm could be
approximated using a series of convolution (computing value function expectation
over stochastic transitions) and maxpooling (choosing the best action) allowing
automatic differentiation.
Okada~et~al.~\cite{Okada_PIN17} proposed path integral networks in which the
control sequence resulting from path integral control may be differentiated with
respect to the controller parameters.
All of these works, however, assume a known environment and only optimize the parameters of a cost function defined over it.

A series of recent works~\cite{ChoiIRLinPOMDPJMLR2011,Shankar_RLviaRCNN2016,
  Wulfmeier_DeepMaxEnt16, Gupta_CMP17, Karkus_QMDPNet2017, Karkus_DAN19} address
IRL under partial observability.
Wulfmeier~et~al.~\cite{Wulfmeier_DeepMaxEnt16} train deep
neural network representations for Ziebart~et~al.'s Maximum Entropy
IRL~\cite{Ziebart_MaxEnt08} method and consider streaming lidar scan
observations of the environment.
Karkus~et~al.~\cite{Karkus_DAN19} formulate the IRL problem as a
POMDP~\cite{ASTROM_POMDP1965}, including the robot pose and environment map in
the state.
Since a na\"ive representation of the occupancy distribution may require
exponential memory in the map size, the authors assume partial knowledge of the
environment (the structure of a building is known but the furniture placement is
not).
A control policy is obtained via the SARSOP~\cite{Kurniawati_SARSOP08} algorithm, which approximates the cost-to-go function only over an optimally reachable space.
Gupta~et~al.~\cite{Gupta_CMP17} address visual navigation in partially observed
environments while using hierarchical VIN as the planner.
Khan~et~al.~\cite{Khan_MACN18} introduce a memory module to VIN to address
partial observability.

Many IRL algorithms rely on dynamic programming, including
 VIN and derivatives~\cite{Gupta_CMP17,Khan_MACN18}, which requires
updating cost-to-go estimates over \emph{all} possible
states.
Our insight is that in partially observable environments, the cost-to-go
estimates need to be updated and differentiated only over a \emph{subset} of
states.
Inspired by~\cite{Ratliff_06}, we obtain cost-to-go estimates only over
promising states using a motion planning algorithm.
This helps to obtain a closed-form subgradient of the
cost-to-go with respect to the learned cost function from the
optimal trajectory.
While Ratiff~et~al.~\cite{Ratliff_06} exploit this observation in fully
observable environments, none of the works focusing on partial observability
take advantage of this.
Our work differs from closely related works in Table~\ref{tab:related-work}.
In summary, we offer two \textbf{contributions} illustrated in Fig.~\ref{fig:approach}:
Firstly, we develop a non-stationary cost function representation composed of a probabilistic occupancy \emph{map encoder}, with recurrent dependence on the observation sequence, and a \emph{cost encoder}, defined over the occupancy features (Sec.~\ref{sec:cost_representation}).
Secondly, we optimize the cost parameters using a closed-form subgradient of the cost-to-go obtained only over a subset of promising states (Sec.~\ref{sec:cost_learning}).

%
\begin{table}
  \scriptsize
  \centering
  \begin{tabular}{lllll}
    \toprule
    $\downarrow$Reference \textbackslash Feature $\rightarrow$ & POE & PBB & DNN & CfG 
    \\
    \midrule
    VIN~\cite{Tamar_VIN16} & \xmark & \xmark & \cmark & \xmark\\
    CMP~\cite{Gupta_CMP17} & \cmark & \xmark & \cmark & \xmark\\
    MaxEntIRL~\cite{Ziebart_MaxEnt08} & \xmark & \xmark & \xmark & \cmark
    \\
    MaxMarginIRL~\cite{Ratliff_06} & \xmark & \cmark & \xmark & \cmark
    \\
    DAN~\cite{Karkus_DAN19} & \cmark$^\dagger$ & \xmark & \cmark & \xmark
    \\
    Ours & \cmark & \cmark & \cmark & \cmark
    \\
    \bottomrule
  \end{tabular}
  \caption{Comparison with closely related work based on the use of Partially Observable Environments (POE),
    Partial Bellman Backups (PBB), Deep Neural Network (DNN) representation, and Closed-form Gradients (CfG). PBB refers to computing and differentiating values over a subset of promising states as opposed to the whole state space. $^\dagger$ DAN~\cite{Karkus_DAN19} works with uncertainty only on the robot or furniture location, while the main environment structure is known.}
    
    
  \label{tab:related-work}
\end{table}

\section{Problem Formulation}
\label{sec:problem_formulation}
Consider a robot navigating in an unknown environment with the task of reaching
a goal state $\bfx_g \in \mathcal{X}$.
Let $\bfx_t \in \mathcal{X}$ be the robot state, capturing its pose, twist,
etc., at discrete time $t$.
For a given control input $\bfu_t \in \mathcal{U}$ where $\calU$ is assumed
finite, the robot state evolves according to known deterministic dynamics:
$\bfx_{t+1} = f(\bfx_t, \bfu_t)$.
Let $m^* : \mathcal{X} \rightarrow \{-1,1\}$ be a function specifying the
\textit{true} occupancy of the environment by labeling states as either feasible
($-1$) or infeasible ($1$) and let $\mathcal{M}$ be the space of possible
environment realizations $m^*$.
Let $c^* : \mathcal{X} \times \mathcal{U} \times \mathcal{M} \rightarrow
\mathbb{R}_{\ge 0}$ be a cost function specifying desirable robot behavior in a
given environment, e.g., according to an expert user or an optimal design.
We assume that the robot does not have access to either the true occupancy map
$m^*$ or the true cost function $c^*$.
However, the robot is able to make observations $\bfz_t \in \mathcal{Z}$ (e.g.,
using a lidar scanner or a depth camera) of the environment in its vicinity,
whose distribution depends on the robot state $\bfx_t$ and the environment
$m^*$.
Given a training set $\mathcal{D} := \crl{(\bfx_{t,n},\bfu_{t,n}^*,\bfz_{t,n}, \bfx_{g,n})}_{t=1, n=1}^{T_n, N}$ of $N$ expert trajectories with length $T_n$ to demonstrate desirable behavior, our goal is to 
\begin{itemize}
    \item learn a cost function estimate $c_t: \mathcal{X} \times \mathcal{U} \times \mathcal{Z}^t \times \Theta \rightarrow \mathbb{R}_{\ge 0}$ that depends on an observation sequence $\bfz_{1:t}$ from the true latent environment and is parameterized by $\bftheta$,
    \item derive a stochastic policy $\pi_t$ from $c_t$ such that the robot behavior under $\pi_t$ matches the prior experience $\mathcal{D}$.
\end{itemize}
To balance exploration in partially observable environments with exploitation of promising controls, we specify $\pi_t$ as a Boltzmann policy~\cite{Ramachandran_BayesianIRL07, Neu_Apprenticeship12} associated with the cost $c_t$:
\begin{equation}
\label{eq:stochastic_policy}
\pi_{t}(\bfu_{t}|\bfx_{t}; \bfz_{1:t}, \bftheta) =
\frac{\exp(-Q^*_{t}(\bfx_{t}, \bfu_{t};\bfz_{1:t}, \bftheta))}
{\sum_{\bfu \in \mathcal{U}}\exp(-Q^*_{t}(\bfx_{t}, \bfu; \bfz_{1:t}, \bftheta))},
\end{equation}
where the optimal cost-to-go function $Q^*_t$ is:
\begin{align}
\label{eqn:Q}
Q^*_t(\bfx_t, \bfu_t ; \bfz_{1:t}, \bftheta)  := \min_{\bfu_{t+1:T-1}} & \sum_{k=t}^{T-1} c_t(\bfx_k, \bfu_k ; \bfz_{1:t}, \bftheta) \\
\text{s.t.}\quad & \bfx_{k+1} = f(\bfx_k, \bfu_k),\; \bfx_T = \bfx_g. \notag
\end{align}

\begin{problem*}
Given demonstrations $\mathcal{D}$, optimize the cost function parameters $\bftheta$ so that log-likelihood of the demonstrated controls $\bfu_{t,n}^*$ is maximized under the robot policies $\pi_{t,n}$:
\begin{equation}
\label{eq:loss}
\min_{\bftheta} \; \mathcal{L}(\bftheta) := - \sum_{n=1}^N \sum_{t=1}^{T_n}\log \pi_{t, n} (\bfu_{t,n}^*|\bfx_{t,n};\bfz_{1:t}, \bftheta).
\end{equation}
\end{problem*}

The problem setup is illustrated in Fig.~\ref{fig:approach}. While Eqn.~\eqref{eqn:Q} is a standard deterministic shortest path (DSP) problem, the challenge is to make it differentiable with respect to $\bftheta$. This is needed to propagate the loss in~\eqref{eq:loss} back through the DSP problem to update the cost parameters $\bftheta$. Once the parameters are optimized, the robot can generalize to navigation tasks in new partially observable environments by evaluating the cost $c_t$ based on the observations $\bfz_{1:t}$ iteratively and (re)computing the associated policy $\pi_t$.


\section{Cost Function Representation}
\label{sec:cost_representation}
\newcommand{\tm}{\bfpsi}
\newcommand{\tmk}{\tm_k}
\newcommand{\tc}{\bfphi}

\begin{figure}[t]
  \centering
  \def\svgwidth{0.9\linewidth}
  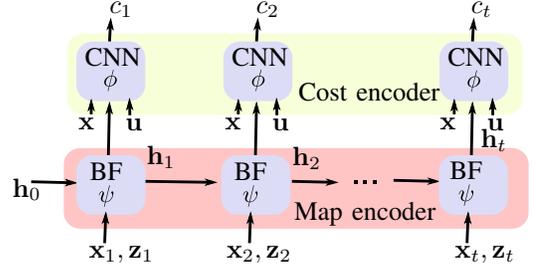
  \caption{Neural network model of a cost function representation. A Bayes filter with likelihood function parameterized by $\tm$, takes in sequential observations $\bfz_{1:t}$ and outputs a latent map representation $\bfh_t$. A convolutional neural network, parameterized by $\tc$, extracts features from the map state to specify the cost $c_t$ at a given robot state-control pair $(\bfx,\bfu)$. The learnable parameters are $\bftheta = \crl{\tm, \tc}$.}
  \label{fig:cost_neural_net}
\end{figure}

We propose a cost function representation comprised of two components: 
a \emph{map encoder} and a \emph{cost encoder}.

The map encoder incrementally updates a hidden state $\bfh_t$ using the most 
recent observation $\bfz_t$ obtained from robot state $\bfx_t$. 
For example, a Bayes filter with likelihood function parameterized by $\tm$ can convert the 
sequential input $(\bfx_{1:t},\bfz_{1:t})$ into a fixed-sized hidden state $\bfh_{t+1}$:
\begin{equation}
\bfh_{t+1} = \textbf{BF}(\bfh_t, \bfx_t, \bfz_t; \tm).
\end{equation}

The cost encoder uses the latent environment map $\bfh_t$ to define the cost function estimate $c_t(\bfx,\bfu)$ at a given state-control pair $(\bfx,\bfu)$. A convolutional neural network (CNN)~\cite{Goodfellow-et-al-2016} with parameters $\tc$ can extract cost features from the environment map:
\begin{equation}
c_t(\bfx,\bfu) = \textbf{CNN}(\bfh_t, \bfx, \bfu; \tc).
\end{equation}

This conceptual model, combining recurrent estimation of a hidden environment state, followed by feature extraction to define the cost at $(\bfx, \bfu)$ is illustrated in Fig.~\ref{fig:cost_neural_net}. The model is differentiable by design, allowing its parameters $\bftheta = \crl{\tm,\tc}$ to be optimized. We propose an instantiation of this general model, specific to modeling occupancy costs from depth measurements in robot navigation tasks.

\subsection{Map Encoder}
\begin{figure}[t]
\centering
\includegraphics[width=0.92\linewidth]{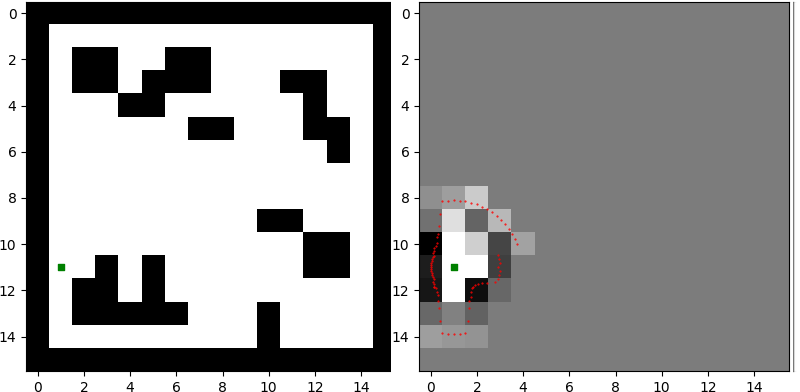}
\caption{A robot (green) navigating in a 2-D grid environment. The true environment (left) is a $16\times 16$ grid where black regions are obstacles and white regions are free. On the right, the noisy lidar beams with endpoints in red have maximum range $2.5$. The robot estimates the occupancy probability through an inverse observation model in Eqn~\eqref{eq:inverse_sensor_model} (darker means higher probability of occupancy).}
\label{fig:map_and_obs}
\end{figure}

We encode the occupancy probability of different environment areas into a hidden state $\bfh_t$. In detail, we discretize $\calX$ into $N$ cells and let $\bfm^* \in \{-1,1\}^N$ be the vector of true occupancy values over the cells. Since $\bfm^*$ is unknown to the robot, we maintain the occupancy likelihood $\mathbb{P}(\bfm^* = \mathbf{1} | \bfx_{1:t}, \bfz_{1:t})$ given the history of states $\bfx_{1:t}$ and observations $\bfz_{1:t}$. See Fig.~\ref{fig:map_and_obs} for an example of a depth measurement $\bfz_t$ and associated occupancy likelihood over the map $\bfm^*$. The representation complexity may be simplified significantly if one assumes independence among the map cells $\bfm^*_j$:
\begin{equation}
\mathbb{P}(\bfm^* = \mathbf{1} | \bfx_{1:t}, \bfz_{1:t}) = \prod_{j=1}^N \mathbb{P}(\bfm^*_j = 1 | \bfx_{1:t}, \bfz_{1:t}) \;.
\end{equation}
We use inspiration from occupancy grid mapping~\cite{Thrun_PR05,Hornung_Octomap13} to design recurrent updates for the occupancy probability of each cell $\bfm^*_j$. Since $\bfm^*_j$ is binary, its likelihood update can be simplified by defining the log-odds ratio of occupancy:
\begin{equation}
\bfh_{t,j} \defeq \log \frac{\mathbb{P}(\bfm^*_j = 1 | \bfx_{1:t}, \bfz_{1:t})}{\mathbb{P}(\bfm^*_j = -1 | \bfx_{1:t}, \bfz_{1:t})} \;.
\end{equation}
The recurrent Bayesian update of $\bfh_{t,j}$ is:
\begin{equation}
\bfh_{t+1,j} = \bfh_{t,j} + \log\frac{p(\bfz_{t+1} | \bfm^*_j = 1, \bfx_{t+1})}{p(\bfz_{t+1} | \bfm^*_j = -1, \bfx_{t+1})} \;,
\end{equation}
where the increment is a log-odds observation model. The occupancy posterior can be recovered from the occupancy log-odds ratio $\bfh_{t,j}$ via a sigmoid function:
\begin{equation}
\label{eq:map_encoder_output}
\mathbb{P}(\bfm^*_j = 1 | \bfx_{1:t}, \bfz_{1:t}) = \sigma\prl{\bfh_{t,j}}.
\end{equation}
The sigmoid function satisfies the following properties:
\begin{equation}
\label{eq:sigmoid_properties}
\scaleMathLine[0.88]{\sigma(x) = \frac{1}{1+e^{-x}}, \;\; 1-\sigma(x) = \sigma(-x), \;\; \log \frac{\sigma(x)}{\sigma(-x)} = x}
\end{equation}

To complete the recurrent occupancy model design, we parameterize the log-odds observation model:
\begin{equation}
\scaleMathLine[0.89]{\log\frac{p(\bfz | \bfm^*_j = 1, \bfx)}{p(\bfz | \bfm^*_j = -1, \bfx)} = \underbrace{\log\frac{\mathbb{P}(\bfm^*_j = 1 | \bfz, \bfx)}{\mathbb{P}(\bfm^*_j = -1 | \bfz, \bfx)}}_{\bfg_j(\bfx,\bfz)} - \bfh_{0,j}}
\end{equation}
where Bayes rule was used to represent it in terms of an inverse observation model $\bfg_j(\bfx,\bfz)$ and a prior occupancy log-odds ratio $\bfh_{0,j}$, whose value may depend on the environment (e.g., $\bfh_{0,j}=0$ specifies a uniform prior over occupied and free cells). Note that map cells $\bfm_j^*$ outside of the sensor field of view at time $t$ are not affected by $\bfz_t$, in which case $\bfg_j(\bfx_t,\bfz_t) = \bfh_{0,j}$. Now, consider a cell $\bfm_j^*$ along the direction of the $k$-th sensor ray, whose depth measurement is $\bfz_{t,k}$. Let $d(\bfx_t,\bfm_j^*)$ be the distance between the robot position and the center of mass of cell $\bfm_j^*$. We model the occupancy likelihood of $\bfm_j^*$ as a truncated sigmoid around the true distance $d(\bfx_t,\bfm_j^*)$:
\begin{equation}
\mathbb{P}(\bfm_j^* = 1 | \bfx_t, \bfz_{t,k}) = \begin{cases}
\sigma( \tmk \delta\bfz_{t,k}) & \text{if } \delta\bfz_{t,k} \leq\epsilon\\
\sigma( \bfh_{0,j} ) & \text{if } \delta\bfz_{t,k} > \epsilon
\end{cases} \;,
\end{equation}
where $\tmk$ is a learnable parameter, $\delta\bfz_{t,k} := d(\bfx_t,\bfm_j^*) - \bfz_{t,k}$, and $\epsilon$ is a distance threshold on the influence of the $k$-th sensor ray on the occupancy of cells around the point of reflection. Using~\eqref{eq:sigmoid_properties}, this implies the following log-odds inverse sensor model for cells $\bfm_j^*$ along the $k$-th sensor ray:
\begin{equation}
\label{eq:simple_inverse_sensor_model}
\bfg_j(\bfx_t,\bfz_t) = \begin{cases}
\tmk \delta\bfz_{t,k} & \text{if } \delta\bfz_{t,k} \leq\epsilon\\
\bfh_{0,j} & \text{otherwise}
\end{cases} \;,
\end{equation}
This model suggests that one may also use a more expressive multi-layer neural network in place of the linear transformation $\tmk \delta\bfz_{t,k}$ of the distance differential along the $k$-th ray:
\begin{equation}
\label{eq:inverse_sensor_model}
\scaleMathLine[0.89]{\bfg_j(\bfx_t,\bfz_t) = \begin{cases}
\textbf{NN}(\bfz_{t,k}, d(\bfx_t, \bfm_j^*); \tmk) & \text{if } \delta\bfz_{t,k} \leq\epsilon\\
\bfh_{0,j} & \text{otherwise}
\end{cases}}
\end{equation}

In summary, the map encoder starts with prior occupancy log-odds $\bfh_{0}$, updates them recurrently via:
\begin{equation}
\bfh_{t+1} = \bfh_{t} + \bfg(\bfx_t,\bfz_t; \tm) - \bfh_{0},
\end{equation} 
where the log-odds inverse sensor model $\bfg_j(\bfx_t,\bfz_t ;\tmk)$ is specified for the $j$-th cell along the $k$-th ray in~\eqref{eq:inverse_sensor_model}, and provides the cell occupancy likelihood in~\eqref{eq:map_encoder_output} as output.



\subsection{Cost Encoder}
Given a state-control pair $(\bfx,\bfu)$, the cost encoder uses the output $\sigma\prl{\bfh_t}$ of the map encoder to obtain a cost function estimate $c_t(\bfx,\bfu)$. A complex interacting structure over the feature representation $\sigma\prl{\bfh_t}$ can be obtained via a deep neural network with parameters $\tc$. Wulfmeier et al.~\cite{Wulfmeier_DeepMaxEnt16} proposed several CNN architectures, including pooling and residual connections of fully convolutional networks, to model $c_t(\bfx,\bfu)$ from $\sigma\prl{\bfh_t}$. While complex architecture may provide better performance in practice, we also develop a simpler model for comparison using the inductive bias of obstacle avoidance in robot navigation.

Let $s$ and $l$ be a small and large positive parameter, respectively. The cost of applying control $\bfu$ in robot state $\bfx$ can be modeled as large when the transition $f(\bfx,\bfu)$ encounters an obstacle and as small otherwise:
\begin{equation*}
c(\bfx,\bfu) := \begin{cases}
s &\text{if } m^*(\bfx) = -1 \text{ and } m^*(f(\bfx,\bfu)) = -1\\
l &\text{if } m^*(\bfx) = 1 \text{ or } m^*(f(\bfx,\bfu)) = 1
\end{cases}
\end{equation*}
Since $m^*$ is unknown, we use the estimated occupancy probabilities $\sigma\prl{\bfh_t}$ to compute the expectation of $c(\bfx,\bfu)$ over $m^*$:
\begin{equation}
\label{eq:cost_model}
\begin{aligned}
c_t(\bfx, \bfu) := & \;\mathbb{E}[c(\bfx, \bfu)] = s\;\sigma\prl{\bfh_t[\bfx]}\sigma\prl{\bfh_t[f(\bfx, \bfu)]}\\
& + l\prl{1-\sigma\prl{\bfh_t[\bfx]}\sigma\prl{\bfh_t[f(\bfx, \bfu)]}},
\end{aligned}
\end{equation}
where $\bfh_t[\bfx]$ is the entry of the map encoder state that corresponds to the environment cell containing $\bfx$. This simple cost encoder has parameters $\tc := [s,\;l]^T$ and its output $c_t(\bfx, \bfu)$ is differentiable with respect to $\tc$ and $\tm$ through $\sigma\prl{\bfh_t}$.

\section{Cost Learning via Differentiable Planning}
\label{sec:cost_learning}


We focus on optimizing the parameters $\bftheta$ of the cost representation $c_t(\bfx,\bfu ; \bfz_{1:t}, \bftheta)$ developed in Sec.~\ref{sec:cost_representation}. Since the true cost $c^*$ is not directly observable, we need to differentiate the loss function $\calL(\bftheta)$ in~\eqref{eq:loss}, which, in turn, requires differentiating through the DSP problem in~\eqref{eqn:Q} with respect to the cost function estimate $c_t$. 

Value Iteration Networks (VIN)~\cite{Tamar_VIN16} shows that $T$ iterations of the value iteration algorithm can be approximated by a neural network with $T$ convolutional and minpooling layers. This allows VIN to be differentiable with respect to the stage cost. While VIN can be modified to operate with a finite horizon and produce a non-stationary policy, it would still be based on full Bellman backups (convolutions and min-pooling) over the entire state space. As a result, VIN scales poorly with the state-space size, while it might not even be necessary to determine the optimal cost-to-go $Q_t^*(\bfx,\bfu)$ at every state $\bfx \in \calX$ and control $\bfu \in \calU$ in the case of partially observable environments.

Instead of using dynamic programming to solve the DSP~\eqref{eqn:Q}, any motion
planning algorithm (e.g., A*~\cite{ARA},
RRT~\cite{Lavalle_RRT98,Karaman_RRTstar11}, etc.) that returns the optimal
cost-to-go $Q_t^*(\bfx,\bfu)$ over a subset of the state-control space provides
an accurate enough solution.
For example, a backwards A* search applied to problem~\eqref{eqn:Q} with start state $\bfx_g$, goal state $\bfx \in \calX$, and predecessors expansions according to the motion model $f$ provides an upper bound to the optimal cost-to-go:
\begin{equation*}
\begin{aligned}
Q_t^*(\bfx,\bfu) &= c_t(\bfx,\bfu) + g(f(\bfx,\bfu)) \; &\forall f(\bfx,\bfu) &\in \text{CLOSED},\\
Q_t^*(\bfx,\bfu) &\leq c_t(\bfx,\bfu) + g(f(\bfx,\bfu)) \; &\forall f(\bfx,\bfu) &\not\in \text{CLOSED},
\end{aligned}
\end{equation*}
where $g$ are the values computed by A* for expanded nodes in the CLOSED list and visited nodes in the OPEN list. Thus, a Boltzmann policy $\pi_t(\bfu \mid \bfx)$ can be defined using the $g$-values returned by A* for all $\bfx \in \text{CLOSED} \cup \text{OPEN} \subseteq \calX$ and a uniform distribution over $\calU$ for all other states $\bfx$. A* would significantly improves the efficiency of VIN~\cite{Tamar_VIN16} or other full backup Dynamic Programming variants by performing local Bellman backups on promising states (the CLOSED list).

In addition to improving the efficiency of the forward computation of $Q_t^*(\bfx,\bfu)$, using a planning algorithm to solve~\eqref{eqn:Q} is also more efficient in back-propagating errors with respect to $\bftheta$. In detail, using the subgradient method~\cite{Shor_Subgradient12,Ratliff_06} to optimize $\calL(\bftheta)$ leads to a closed-form (sub)gradient of $Q_t^*(\bfx_t,\bfu_t)$ with respect to $c_t(\bfx,\bfu)$, removing the need for back-propagation through multiple convolutional or min-pooling layers. We proceed by rewriting $Q_t^*(\bfx_t,\bfu_t)$ in a form that makes its subgradient with respect to $c_t(\bfx,\bfu)$ obvious. Let $\calT(\bfx_t,\bfu_t)$ be the set of feasible state-control trajectories $\bftau := \bfx_t,\bfu_t,\bfx_{t+1},\bfu_{t+1},\ldots,\bfx_{T-1},\bfu_{T-1}$ starting at $\bfx_t$, $\bfu_t$ and satisfying $\bfx_{k+1} = f(\bfx_k,\bfu_k)$ for $k = t,\ldots,T-1$ with $\bfx_g = \bfx_T$. Let $\bftau^* \in \calT(\bfx_t,\bfu_t)$ be an optimal trajectory corresponding to the optimal cost-to-go function $Q^*_t(\bfx_t,\bfu_t)$ of a deterministic shortest path problem, i.e., the controls in $\bftau^*$ satisfy the additional constraint $\bfu_{k} = \arg\min_{\bfu \in \calU} Q^*_t(\bfx_{k},\bfu)$ for $k = t+1,\ldots,T-1$. Let $\mu_{\bftau}(\bfx,\bfu) := \sum_{k=t}^{T-1} \indicator_{(\bfx_k,\bfu_k) = (\bfx,\bfu)}$ be a state-control visitation function indicating if $(\bfx,\bfu)$ is visited by $\bftau$. With these definitions, we can view the optimal cost-to-go function $Q^*_t(\bfx_t,\bfu_t)$ as minimum over $\calT(\bfx_t,\bfu_t)$ of the inner product of the cost function $c_t$ and the visitation function $\mu_{\bftau}$:
\begin{equation}
\label{eq:inner_product_q}
Q^*_t(\bfx_t,\bfu_t) = \min_{\bftau \in \calT(\bfx_t,\bfu_t)}  \sum_{\bfx \in \mathcal{X},\bfu\in\mathcal{U}} c_t(\bfx,\bfu) \mu_{\bftau}(\bfx,\bfu)
\end{equation}
where $\calX$ can be assumed finite because both $T$ and $\calU$ are finite. This form allows us to (sub)differentiate $Q^*_t(\bfx_t,\bfu_t)$ with respect to $c_t(\bfx,\bfu)$ for any $\bfx \in \calX$, $\bfu \in \calU$.

\begin{lemma}
\label{lemma:subgradient}
Let $f(\bfx,\bfy)$ be differentiable and convex in $\bfx$. Then, $\nabla_{\bfx} f(\bfx, \bfy^*)$, where $\bfy^* := \arg\min_{\bfy} f(\bfx,\bfy)$, is a subgradient of the piecewise-differentiable convex function $g(\bfx) := \min_{\bfy} f(\bfx,\bfy)$.
\end{lemma}

Applying Lemma~\ref{lemma:subgradient} to~\eqref{eq:inner_product_q} leads to the following subgradient of the optimal cost-to-go function:
\begin{equation}
\label{eq:subgradient_q}
\frac{\partial Q^*_t(\bfx_t,\bfu_t)}{\partial c_t(\bfx,\bfu)} = \mu_{\bftau^*}(\bfx,\bfu),
\end{equation}
which can be obtained from the optimal trajectory $\bftau^*$ corresponding to $Q^*_t(\bfx_t,\bfu_t)$. This result and the chain rule allow us to obtain the complete (sub)gradient of $\calL(\bftheta)$.

\begin{proposition}
\label{prop:chain_rule}
A subgradient of the loss function $\calL(\bftheta)$ in~\eqref{eq:loss} with respect to the cost function parameters $\bftheta$ can be obtained as follows:
\begin{align}
&\frac{d \calL(\bftheta)}{d \bftheta} = - \sum_{n=1}^N \sum_{t=1}^{T_n} \frac{d \log \pi_{t,n}(\bfu_{t,n}^* \mid \bfx_{t,n})}{d \bftheta}\\
&=- \sum_{n=1}^N \sum_{t=1}^{T_n}\sum_{\bfu_{t,n} \in \calU} \frac{\partial \log \pi_{t,n}(\bfu_{t,n}^* \mid \bfx_{t,n})}{\partial Q_{t,n}^*(\bfx_{t,n},\bfu_{t,n})} \frac{d Q_{t,n}^*(\bfx_{t,n},\bfu_{t,n})}{d \bftheta}\notag
\end{align}
where the first term has a closed-form, while the second term is available from~\eqref{eq:subgradient_q} and the cost representation in Sec.~\ref{sec:cost_representation}:
\begin{align}
&\frac{\partial \log \pi_{t,n}(\bfu_{t,n}^* \mid \bfx_{t,n})}{\partial Q_{t,n}^*(\bfx_{t,n},\bfu_{t,n})} = \prl{\indicator_{\{\bfu_{t,n} = \bfu_{t,n}^*\}} - \pi_{t,n}(\bfu_{t,n} | \bfx_{t,n})}\notag\\
&\frac{d Q_{t,n}^*(\bfx_{t,n},\bfu_{t,n})}{d \bftheta} = \!\!\!\!\!\sum_{(\bfx,\bfu) \in \bftau^*} \!\!\!\!\!\frac{\partial Q_{t,n}^*(\bfx_{t,n},\bfu_{t,n})}{\partial c_t(\bfx,\bfu)} \frac{\partial c_t(\bfx,\bfu)}{\partial \bftheta}
\end{align}
\end{proposition}

The computation graph structure implied by Prop.~\ref{prop:chain_rule} is illustrated in Fig.~\ref{fig:approach}. The graph consists of a cost representation layer and a differentiable planning layer, allowing end-to-end minimization of $\calL(\bftheta)$ via stochastic (sub)gradient descent. Full algorithms for the training and testing phases (Fig.~\ref{fig:approach}) are shown in Alg.~\ref{alg:train} and Alg.~\ref{alg:test}.

Although the form of the gradient in Prop.~\ref{prop:chain_rule} is similar to
that in Ziebart et al.~\cite{Ziebart_MaxEnt08}, our contributions are
orthogonal. Our contribution is to obtain a \emph{non-stationary} cost-to-go
function and its (sub)gradient for a finite-horizon problem using forward and
backward computations that scale efficiently with the size of the state space.
On the other hand, the results of Ziebart et al.~\cite{Ziebart_MaxEnt08} provide
a \emph{stationary} cost-to-go function and its gradient for an infinite horizon
problem.
The maximum entropy formulation of the stationary policy is a well grounded property 
of using a soft version of the Bellman update, which can explicitly model the 
suboptimality of expert trajectories.
However, we can show the benefits of our approach without including the maximum entropy
formulation and will leave it as future work.

\begin{algorithm}[t]
\caption{Training: learn cost function parameters $\bftheta$}
\label{alg:train}
\small
\begin{algorithmic}[1]
\State {\bfseries Input:} Demonstrations $\mathcal{D} \!=\! \crl{(\bfx_{t,n},\bfu_{t,n}^*,\bfz_{t,n}, \bfx_{g,n})}_{t=1, n=1}^{T_n,N}\!\!$
\While{$\bftheta$ not converged}
\State $\mathcal{L}(\bftheta) \gets 0$
\For{$n = 1, \ldots,N$ \textbf{and} $t = 1,\ldots, T_n$}
  \State Update $c_{t,n}$ based on $\bfx_{t,n}$ and $\bfz_{t,n}$ as in Sec.~\ref{sec:cost_representation}
  \State Obtain $Q_{t,n}^*(\bfx,\bfu)$ from DSP~\eqref{eqn:Q} with stage cost $c_{t,n}$
  \State Obtain $\pi_{t,n}(\bfu | \bfx_{t,n})$ from $Q_{t,n}^*(\bfx_{t,n},\bfu)$ via Eq.~\eqref{eq:stochastic_policy}
  \State $\mathcal{L}(\bftheta) \gets \mathcal{L}(\bftheta) -\log \pi_{t, n} (\bfu_{t,n}^*|\bfx_{t,n})$ 
\EndFor
\State Update $\bftheta \gets \bftheta - \alpha \nabla \mathcal{L}(\bftheta)$ via Prop.~\ref{prop:chain_rule}
\EndWhile
\State {\bfseries Output:} $\bftheta$ 
\end{algorithmic}
\end{algorithm}

\begin{algorithm}[t]
\caption{Testing: compute control policy for learned $\bftheta$}
\label{alg:test}
\small
\begin{algorithmic}[1]
\State {\bfseries Input:} Start state $\bfx_s$, goal state $\bfx_g$, optimized $\bftheta$ 
\State Current state $\bfx_t \gets \bfx_s$
\While{$\bfx_t \neq \bfx_g$}
    \State Make an observation $\bfz_t$
    \State Update $c_t$ based on $\bfx_t$ and $\bfz_t$ as in Sec.~\ref{sec:cost_representation}
    \State Obtain $Q_t^*(\bfx_t,\bfu)$, $\bfu \in \calU$ from DSP~\eqref{eqn:Q} with stage cost $c_t$
    \State Obtain $\pi_t(\bfu | \bfx_t)$ from $Q_t^*(\bfx_t,\bfu)$ via Eq.~\eqref{eq:stochastic_policy}
    \State Update $\bfx_{t} \gets f(\bfx_t, \bfu_t)$ via $\bfu_t := \arg\max_\bfu \pi_t(\bfu|\bfx_t)$
\EndWhile
\State {\bfseries Output:} Navigation succeeds or fails.
\end{algorithmic}
\end{algorithm}

\section{Experiments}
\label{sec:experiments}
\label{sec:2d_grid}

\begin{figure*}[t]
      \scriptsize
    \begin{minipage}[c]{0.78\linewidth}
    \begin{tabular}{c c c c c c c c c}
      \toprule
      &  \multicolumn{4}{c}{$16\times 16$ } & \multicolumn{4}{c}{$100\times 100$ } \\
      \cmidrule(r){2-5} \cmidrule(r){6-9}
      Model & \makecell{Val. loss} & \makecell{Val. acc.\\ (\%)} & \makecell{Test traj. \\ succ. rate (\%)} & \makecell{Test \\ traj. diff.}
      & \makecell{Val. loss} & \makecell{Val. acc.\\ (\%)} & \makecell{Test traj. \\ succ. rate (\%)} & \makecell{Test \\ traj. diff.}
      \\
      \midrule
        \texttt{DeepMaxEnt}~\cite{Wulfmeier_DeepMaxEnt16} & 0.18 &  93.6 & 90.9 & 0.145  & 0.23  & 93.7 & 31.1 & 6.528 \\
        \texttt{Ours-HCE} & 1.10 &  59.5 & 99.7 & 0.378  & 1.32  & 42.2 & 100.0 & 2.876\\
        \texttt{Ours-SCE} & 0.66 &  62.1 & 97.2 & 0.174  & 0.66  & 62.1 & 84.4 & 1.569 \\
        \texttt{Ours-CNN} & 0.27 &  90.5 & 96.7 & 0.144  & 0.14  & 95.1 & 90.1 & 1.196 \\
        \bottomrule
    \end{tabular}%
    \end{minipage}%
    \begin{minipage}[c]{0.20\linewidth}
        \includegraphics[trim=0 0 0 0,clip,width=1.1\linewidth]{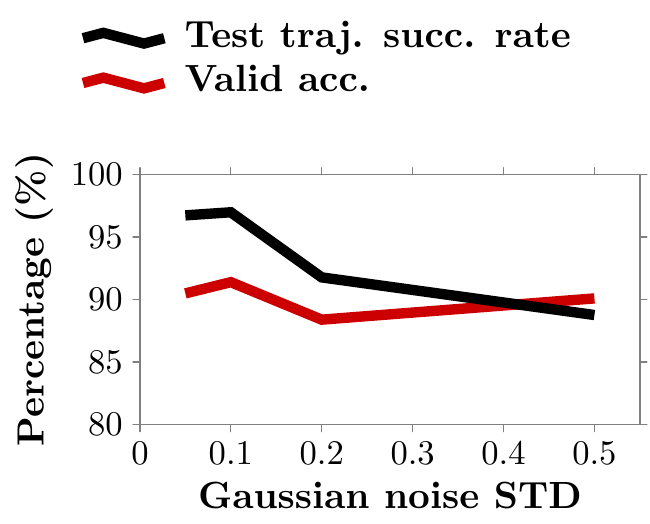}
    \end{minipage}
    \caption{
    Validation and test results for the $16\times16$ and $100\times100$ grid world domains. 
    Cross entropy loss~\eqref{eq:loss} and prediction accuracy for the validation set are reported.
    Test trajectories are iteratively rolled out from the non-stationary policy $\pi_t$.
    A trial is classified as successful if the goal is reached without collisions within twice the number of steps of a shortest path in the groundtruth enviroment. \texttt{Ours-CNN} is capable of matching the expert demonstrations 
    while generalizing to new robot navigation tasks in test time.
    Right: Plot showing the effect of noise on the accuracy of \texttt{Ours-CNN} model.
    }
    \label{tb:test_results}
   \label{fig:map16_noise}
\end{figure*}

\begin{figure*}[t]
  \centering
	\includegraphics[width=0.9\linewidth]{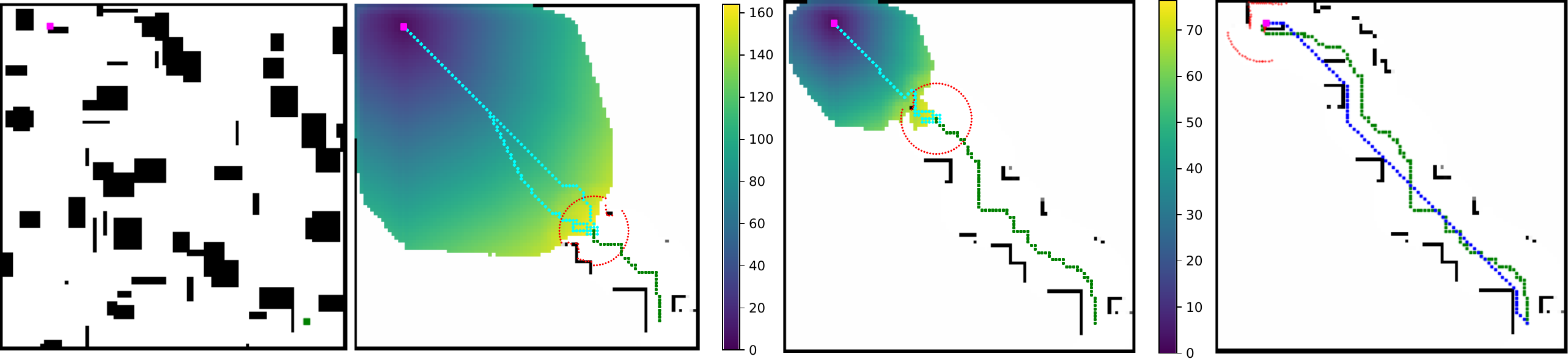}
    \caption{
    Examples of occupancy estimation, A* motion planning and subgradient computation 
    during a successful test trajectory. 
    The first figure shows the \textit{true} occupancy map $\bfm^*$ with the robot start and 
    goal locations in green and magenta, respectively. 
    The second and third figures show the current lidar observation $\bfz_t$ in red and
    robot trajectory thus far $\bfx_{1:t}$ in green.
    The map occupancy estimate $\sigma(\bfh_t)$ in grayscale is in the background.
    The optimal cost-to-go estimate $g$ from A* is shown in a blue-yellow colormap in the foreground (brighter means higher cost-to-go).
    The optimal trajectories $\tau^*$ in cyan corresponding to $Q_t^*(x_t, u_t)$ are obtained 
    during A* planning for subgradient computation in Eqn.~\eqref{eq:subgradient_q}. 
    The last figure shows the final successful trajectory in green and an optimal shortest path in the fully observable enviroment in blue.} 
  \label{fig:test_trajectory}
\end{figure*}

%
%
%


We evaluate our approach in 2D grid world navigation tasks at two scales.
Obstacle configurations are generated randomly in maps $\bfm_n^*$ of sizes
$16\times16$ or $100\times100$. We use an 8-connected grid so that a control
$\bfu_t$ causes a transition $\bfx_{t+1} = f(\bfx_t, \bfu_t)$ from $\bfx_t$ to
one of the eight neighbor cells $\bfx_{t+1}$. At each step, the robot receives a
$360^{\circ}$ lidar scan $\bfz_t$ at $5^{\circ}$ resolution, resulting in $72$
beams $\bfz_{t,k}$ in each scan (see Fig.~\ref{fig:map_and_obs}). The lidar
range readings are corrupted by an additive zero mean Gaussian noise.
The standard deviation of the noise is $0.05$ and $0.2$ (grid cell = 1) and the
lidar maximum range is $2.5$ and $10$ in $16\times16$ and $100\times100$
domains, respectively.
Note that the lidar range is smaller than the 
map size to demonstrate environment partial observability. 
During test time, the domain size is the maximum size allowed for the observed area along
a trajectory. 
Demonstrations are obtained by running an A* planning algorithm to solve the
deterministic shortest path problem on the true map $\bfm^*_n$.
The number of maps and training samples generated are shown in Table~\ref{tab:dataset-size}.

\begin{table}
  \centering
    \begin{tabular}{c c c c c}
      \toprule
      & \multicolumn{2}{c}{$16\times 16$} &
      \multicolumn{2}{c}{$100\times 100$} \\
      \cmidrule(r){2-3} \cmidrule(r){4-5}
      Dataset  & \#maps & \#samples & \#maps & \#samples \\
      \midrule
      Train & 7638 & 514k & 970 & 460k\\
      Validation & 966 & 66k & 122 & 58k \\
      Test & 952 & - & 122 & - \\
      \bottomrule
    \end{tabular}
    \caption{Dataset size. 
    We sample 10 trajectories in each map in training and validation,
    and each sample takes the form $(\bfx_{1:t, n}, \bfu_{t,n}^*, \bfz_{1:t,n}, \bfx_{g,n})$.
    In testing, the robot's task is to navigate from one randomly sampled start to goal location
    on each map.}
    \label{tab:dataset-size}
\end{table}

\subsection{Baseline and model variations}
We use \texttt{DeepMaxEnt}~\cite{Wulfmeier_DeepMaxEnt16} as a baseline and compare it to three variants of our model.
In all variants, we parameterize an inverse observation model and use the log-odds update rule in Eqn.~\eqref{eq:simple_inverse_sensor_model} as the map encoder. This map encoder is sufficiently expressive to model occupancy probability from lidar observations in a 2D enviroment. The A* algorithm is used to solve the DSP~\eqref{eqn:Q}, providing the optimal cost-to-go $Q^*_t(\bfx,\bfu)$ for $\bfx \in \text{CLOSED} \cup \text{OPEN}$ and a subgradient of $Q^*_t(\bfx_t,\bfu_t)$ according to~\eqref{eq:subgradient_q}. All the neural networks are implemented in the PyTorch library~\cite{Paszke_Pytorch17} and trained with the Adam optimizer~\cite{Kingma_ADAM14} until convergence.



\texttt{DeepMaxEnt} uses a neural network to learn a cost function directly from lidar observations 
without explicitly having a map representation. The neural network in our experiments is the ``Standard FCN'' 
in~\cite{Wulfmeier_DeepMaxEnt16} in the $16\times16$ domain, and the encoder-decoder architecture in~\cite{SegNet} in the $100\times100$ domain. Value iteration is approximated by a finite number of Bellman backup iterations, equal to the map size. 
The experiments in the original DeepMaxEnt paper~\cite{Wulfmeier_DeepMaxEnt16} use the mean and variance of the height of the 3D lidar points in each cell, as well as a binary indicator of cell visibility, as input features to the neural network.
Since our synthetic experiments are set up in 2D, the count of lidar beams in each cell is used as replacement of the height mean and variance. This is a fair adaptation because Wulfmeier~et.~al.~\cite{Wulfmeier_DeepMaxEnt16} argued that obstacles generally represent areas of larger height variance which means more beam counts in our observations.

\texttt{Ours-HCE} stands for hard-coded cost encoder. This simple variant of our model uses Eqn.~\eqref{eq:cost_model} with $s=1$ and $l=100$ set explicitly as constants.

\texttt{Ours-SCE} stands for soft-coded cost encoder and has $s$ and $l$ in Eqn~\eqref{eq:cost_model} as learnable parameters. 

\texttt{Ours-CNN} is our most generic variant using a convolutional neural network as cost encoder. The network architecture is the same as in \texttt{DeepMaxEnt} for fair comparison.

\subsection{Experiments and Results}
\paragraph{Model generalization}
Fig~\ref{tb:test_results} shows the comparison of multiple measures of accuracy for different algorithms. 
Both \texttt{Ours-HCE} and \texttt{Ours-SCE} explicitly incorporate cell traversability through the cost design in~\eqref{eq:cost_model}. 
Test results show that this explicit cost encoder is successful at obstacle avoidance, regardless of whether the parameters $s$, $l$ are constants in \texttt{Ours-HCE} or learnable in \texttt{Ours-SCE}. 
The performance of \texttt{Ours-HCE} also shows that the map encoder is learning a correct map representation from the noisy lidar observations since the only trainable parameters are $\tm$ in the inverse observation model~\eqref{eq:simple_inverse_sensor_model}.
However, both models fail at matching demonstrations in validation because the cost encoder~\eqref{eq:cost_model} emphasizes obstacle avoidance explicitly, leaving little capacity to learn from demonstrations. 
\texttt{Ours-CNN} combines the strength of learning from demonstrations and generalization to new navigation tasks while avoiding obstacles. 
Its validation results are on par with \texttt{DeepMaxEnt}, showing the validity of the closed-form subgradient in~\eqref{eq:subgradient_q}. 
\texttt{Ours-CNN} significantly outperforms \texttt{DeepMaxEnt} in new tasks at test time. The performance gap of \texttt{DeepMaxEnt} in the two domains shows that a general CNN architecture applied directly to the lidar scan measurements is not as effective as the map encoder in \texttt{Ours-CNN} at modeling occupancy probability. 
Fig~\ref{fig:test_trajectory} shows the map occupancy estimation, as well as the optimal trajectories necessary for subgradient computation in Sec~\ref{sec:cost_learning}.

\paragraph{Robustness to noise}
The robustness of \texttt{Ours-CNN} to the observation noise is evaluated in the $16\times16$ domain. Fig~\ref{fig:map16_noise} shows that the performance degrades as the noise increases but our inverse observation model~\eqref{eq:simple_inverse_sensor_model} still generalizes well considering that noise with standard deviation of $0.5$ is significant when the lidar range is only $2.5$.

\paragraph{Computational efficiency}
Finally, we compare the efficiency of a forward pass through our A* planner and the value iteration algorithm in \texttt{DeepMaxEnt}. The A* algorithm in our models is implemented in C++ and evaluated on a CPU. The VI algorithm is implemented using convolutional and minpooling layers in Pytorch as described in~\cite{Tamar_VIN16} and is evaluated on a GPU. We record the time that each models takes to return a policy $\pi_t$ given a cost function $c_t$. Our A* algorithm takes only $0.02$ ms as compared to VI's $0.2$ ms on the $16\times 16$ map. In the $100\times 100$ domain, our A* algorithm takes $0.6$ ms, compared to VI's $14$ ms, illustrating the scalability of our model in the size of the state space.





\section{Conclusion}
\label{sec:conclusion}

We proposed an inverse reinforcement learning approach for infering navigation costs from demonstration in partially observable enviroments. Our model introduces a new cost representation composed of a probabilistic occupancy encoder and a cost encoder defined over the occupancy features. We showed that a motion planning algorithm can compute optimal cost-to-go values over the cost representation, while the cost-to-go (sub)gradient may be obtained in closed-form. Our work offers a promising model for encoding occupancy features in navigation tasks and may enable efficient online learning in challenging operational conditions.



\newpage
{\small
\bibliographystyle{cls/IEEEtran}
\bibliography{bib/ref.bib}
}



\end{document}